\documentclass[runningheads,a4paper]{llncs}
\usepackage{graphicx}
\usepackage{cite}
\usepackage{amsmath}
\usepackage{amsfonts}
\usepackage{subfig}
\usepackage{makecell}

\title{Automated Text Mining of Experimental Methodologies from Biomedical Literature}

\titlerunning{CSC8499: Individual Project}

\author{Ziqing Guo}

\institute{MSc in Advanced Computer Science, \\ School of Computing Science,
        University of Newcastle, U. K. \\
        \email{
Z.Guo25@newcastle.ac.uk}
}

\begin{document}

\maketitle 

\begin{abstract}
Biomedical literature is a rapidly expanding field of science and technology. Classification of biomedical texts is an essential part of biomedicine research, especially in the field of biology. This work proposes the fine-tuned DistilBERT, a methodology-specific, pre-trained generative classification language model for mining biomedicine texts. The model has proven its effectiveness in linguistic understanding capabilities and has reduced the size of BERT models by 40\% but by 60\% faster. The main objective of this project is to improve the model and assess the performance of the model compared to the non-fine-tuned model. We used DistilBert as a support model and pre-trained on a corpus of 32,000 abstracts and complete text articles; our results were impressive and surpassed those of traditional literature classification methods by using RNN or LSTM. Our aim is to integrate this highly specialised and specific model into different research industries.
\end{abstract}

\section{Introduction}
Interest in the text extraction of biomedical literature has been growing since the amount of biomedical literature is expanding at a rapid pace. Since 1996, at least 64 million academic papers have been published, with the number of new publications steadily growing.\cite{wordsrate2023}. As of May 2023, according to PubMed, approximately 35\footnote{https://www.ncbi.nlm.nih.gov/pmc/about/intro/} million academic articles have been published, including short surveys, reviews, and conference proceedings. According to the information, about 3,000 new articles were published in peer-reviewed scholarly journals in English, omitting preprints and technical reports, such as clinical trial reports, in various archives\cite{Lee_2019}. Thus, biomedical literature text mining has become an essential requirement.

Nevertheless, there are constraints to directly applying avant-garde natural language processing (NLP) approaches to biomedical text mining. To begin, because contemporary word representation models (LMS) such as Context2Vec\cite{Melamud_2016}, ELMo\cite{Peng_2019}, CoVe\cite{Pajon_2021}, GloVe\cite{Pennington_2014}, and FastText\cite{Liao_2016} are primarily trained and evaluated on datasets comprising general domain texts, evaluating the efficacy of recent word representation models on datasets comprised biomedical texts poses a significant challenge. Moreover, the dissimilarities in word distributions between known public corpora and biomedical databases can pose challenges for biomedical text mining models\cite{Jaderberg_2016}. In addition to the disparity of the different corpus, the aforementioned factors make text mining and representation learning more difficult. Text analysis requires more than just understanding semantics and syntax; understanding context is equally crucial\cite{Naseem_2020}.

In recent progress, pre-training models such as GPT-3 \cite{Liu_2021} and BERT are effective in natural language processing (NLP). Previously, large language models (LLM)\footnote{https://www.elastic.co/what-is/large-language-models} were trained and fine-tuned for various tasks. Nowadays, the GPT-like model may generalise to previously unforeseen scenarios with a few in-context examples, offering up numerous new technical possibilities previously thought to be exclusive to humans (autoregressive language modelling). As a result, this model is frequently utilised in language-generating jobs (e.g. bullet point creation and auto-reply robots). However, BERT-related model mainly focuses on sequence classification or semantic understanding tasks. The GLUE benchmark\cite{Wang_2018}, which includes nine language comprehension problems that span a wide range of domains, facilitates researchers in analysing the NLP models. By analysing the leaderboard (https://gluebenchmark.com/leaderboard/), especially in the classification tasks, BERT models have shown their ability to perform classification tasks, but when applied directly to biological tasks, their performance is deficient.

This work proposes the fine-tuned DistilBERT \cite{Sanh_2019}, a methodology-specific, pre-trained generative classification language model for the mining of biomedical texts. Previously, BioBert\cite{Lee_2019} and BioGPT \cite{Luo_2022}, two of the most prominent context-dependent word presentation models, were trained in the vast biomedical corpus, which contained billions of words and phrases that were not specific for the classification of the article methodology. Contextualised word representations have demonstrated their effectiveness; however, their performance in biomedical corpora is unsatisfactory as a result of their sole training in general domain corpora\cite{Cohen_2013}. Most of the previous work explored distillation to construct task-specific models\cite{Cohen_2013}. However, the DistilBERT pretraining model has proven its effectiveness in linguistic understanding capabilities and reduced the size of the BERT model by 40\% but 60\% faster. To better simulate article methodology classification tasks, we carefully constructed more than 20,000 corpora labelled by different methods extracted from ontology for fine-tuning the DistilBERT\cite{Sanh_2019} model. The results showed that the fine-tuned version of the DistilBERT\cite{Sanh_2019} model achieved low evaluation and training loss.

\subsection{Aim and Objectives}

The primary aim of this project is to fine-tune the model, making it designed explicitly for sequence classification tasks - biomedical literature methodology classification, and evaluate the comparison between the model's performance and that of the not fine-tuned model.

\noindent The following are the main objectives of the project:

\begin{enumerate}
    \item \textit{Review GPT-related and transformer mechanics and investigate the performance of the BERT-like model in document classification and its downstream projects.} After conducting a thorough investigation of various corpora of articles on natural language processing, analysing and comparing the advantages and disadvantages of the model makes the final choice of model more accurate. 
    
    \item \textit{Extract all laboratory techniques and experimental design terms from the NCBO ontology\cite{Martínez-Romero_2016}} Our focus was on methodology techniques to extract valuable insights from the articles. Ontology Recommender 2.0(NCBO) recommends appropriate ontologies when annotating biomedical text data. This approach requires the model to understand the various strategies used in the biomedical field and to identify the most related techniques to achieve accurate and reliable results. 
    
    \item \textit{Construct the streaming pipeline to retrieve the abstracts, methods, and results of the abstracts, methods, and results.} Based on extracted terms, the constructed dictionary includes both full-text and multiple labels of articles. Various pre-processed reshapes into a structured and machine-readable sequence of binary matrix illustrate the cluster proximity by the labelled dendrogram.
    
    \item \textit{Feed the preprocessed data into the model and use a prompt-based strategy to retrain the model.} One of the model's objectives is to self-teach to distinguish which papers belong to which cluster and which terms related to the methodology are contained. This makes data classification more efficient and accurate, allowing valuable information to be obtained and meaningful conclusions to be drawn. 
    
    \item \textit{Evaluate the results of the methodology classification in unseen articles} The model analyses a sample of articles that have not yet been previously classified (raw text) and generates logits\footnote{Before applying an activation function like softmax, a model produces raw output values known as logits. These values are frequently used as input for other functions, such as cross-entropy loss.} based on multiple methodology classification criteria. This can provide insight into reliability and consistency and help researchers quickly identify which methods have been used in the field.
\end{enumerate}

\subsection{Structure of Dissertation}
The structure of this paper is as follows: 
\begin{itemize}
    \item \textit{Background Research}
    This section explores transformer architecture, biomedical text extraction strategies and current challenges and opportunities. In addition, it reviews the current state-of-the-art model and previous methods and approaches proposed or implemented to solve the same or similar problems. 
    \item \textit{Methods}
    This section explains the methods used to investigate a multi-label sequence classification problem, data acquisition, and data preprocessing and visualisation. Also, illustrate the data flow pipeline architecture, including fine-tuning the model and parameter adjustment.
    \item \textit{Results and Discussion}
    In this section, we elaborate on the challenges of data flow implementation and evaluate the performance of the model in classifying the methodology of the biomedical literature. This article discusses the project's extension downstream and compares our fine-tuned model with recent state-of-the-art models.
    \item \textit{Conclusion}
    This section recaps the finished work, re-evaluates the goals and objectives, and proposes potential usability for the research worker and future work.
\end{itemize}

\section{Background Research}

\subsection{Related Work}

The volume of biomedical literature is expanding in response to recent public health events, and manual cataloguing methods are under extreme strain \cite{Koutsomitropoulos_2021}. For example, according to a recent study, the average acceptance period for COVID-19-related articles was six days. The average acceptance period for articles related to Ebola was 15 days and 102 days, respectively, although these are unpopular topics\cite{Chen_2020}. This phenomenon was reflected in the database of hot topics in the biomedical literature, the latest SARS-CoV-2 and COVID-19, which proliferated with approximately 10,000 new articles added monthly. 

The present word embedding algorithm, according to Andriopoulos and colleagues, can efficiently enable biomedical text categorisation in multiple-label contexts\cite{Koutsomitropoulos_2021}. They report and display the F1 scores of each model at each cardinality. Manual indexing of the biological and medical literature is a time-consuming and challenging operation. The team compared pre-trained and transfer learning models for multilabel classification using structured semantic representations in Web Ontology Language (OWL).\cite{McGuinness_2004}.

Mikolov and his colleagues compared the findings of various models, including the vector space word model\cite{le2014distributed}. Studies produce F scores ranging from 0.34 to 0.77, depending on the model tested and the total number of classes. These results are better than or equivalent to the present state of the art, owing to the minimal number of classes utilised in current benchmarks for multi-label classification. An example is that the GLUE comprises a corpus of language acceptance (COLA) files\cite{GLUE2023}. The next objective is to divide each statement into two categories based on whether it is a grammatical English sentence. When considering the Medical Subject Headings(MeSH), the classification score ranges between 0.61 and 0.69\cite{Mao_2017, You_2020}. As a result, transfer learning via pre-trained models is preferable, and an extension of a fully connected classification layer to a pre-trained model can significantly enhance classification accuracy. 

DeepMeSH \cite{You_2020}uses deep semantic information to create large-scale MeSH indexes. It tackles both the citation and classification sides of the problem. A new deep semantic representation solves the citation side difficulty of combining sparse and dense semantic representations, D2V-TFIDF. MeSHLabeler's "learning-to-rank" architecture, which incorporates various evidence generated from the new semantic representation and achieves a micro-F measurement of 0.6323, solves the MeSH term side difficulty\cite{Mao_2017}.

BioWordVec\cite{Zhang_2019} is a set of biomedical word embeddings that incorporate subword information from unlabelled biomedical texts with a biomedical control vocabulary named MeSH. It was evaluated on several BioNLP tasks, such as similarity sentence pairs, protein-protein interaction extraction, and drug-pharmaceutical interaction extraction, with an increase in the F-point from 0.687 to 0.724.

LitMC-BERT\cite{Qingyu_2022} is a transformer-based multi-label classification model for biomedical literature, which employs BioBERT as the foundation model and adds two modules: a label module and a label pair module. The model was evaluated in two sets: LitCovid BioCreative and HoC (cancer hallmarks) and compared to ML-NET, binary BERT, and linear BERT. The results showed that the instance-based F1 and accuracy were 3 and 8\% higher than BERT baseline. 

\subsection{Natural Language Processing}
\begin{figure}[!htb]
\center
\includegraphics[scale=0.5]{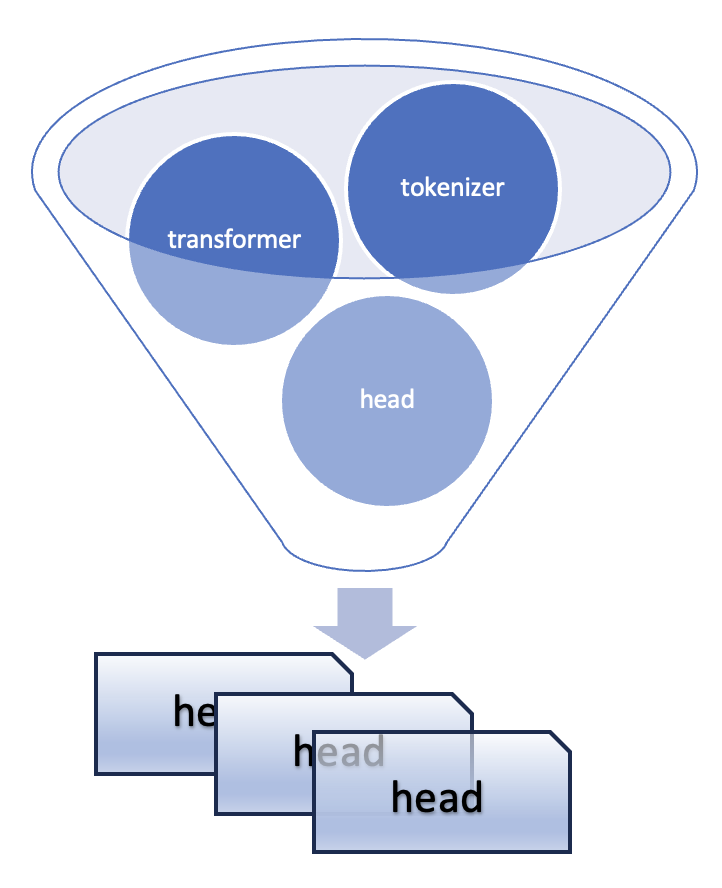}
\caption{Transformers, tokenizers and head make each model, and the mechanics of generated output is such as a funnel.}
\label{firstFig}
\end{figure}

\begin{figure}[!htb]
\center
\includegraphics[scale=0.14]{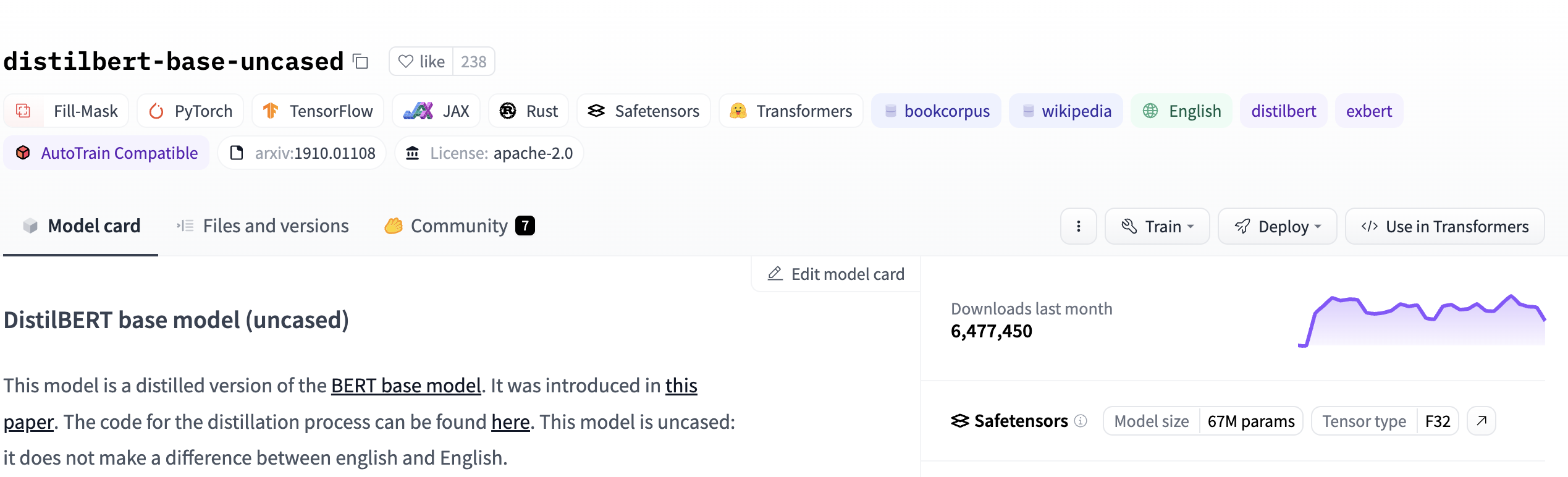}
\caption{(Left) Example of a model page and model card for DistilBERT\cite{Sanh_2019} base model, a smaller and faster transformer model than BERT, based on BERT model as a backbone model, using the same body as self-supervision. (Right) The statistic graph of inference times for the pre-trained BERT model is shown above, with model size and tensor type below. The data science engineer can use this model with a pipeline for modelling the mask language.}
\label{secondFig}
\end{figure}

Natural language processing (classification of text) aims to predict one or more labels for a particular text. It is assumed that each document is labelled with a class in traditional classification tasks\cite{Boutell_2004}. On the other hand, multilabel classification implies that documents can be assigned simultaneously and independently to multiple labels or classes. Huggingface \cite{Wolf_2019} provides a state-of-the-art transformer toolkit for standard natural language processing pipelines. The following three elements are defined for each model in the HuggingFace transformer library and are shown in Fig.\ref{firstFig}: (i) tokenising the raw text, (ii) converting the low-index encoding, (iii) converting the low-index into the context index, and determining the task-specific predictions using the context index. Most of the text classification can be addressed with these three components. There are various transformer models that can be used for various NLP applications. These models are specifically designed for tasks such as understanding, generation, and conditional generation. In addition, models are created for special applications, such as quick conclusions or multilingual applications. Tokenizers can implement additional valuable features for users and incorporate additional beneficial features for consumers. It ranges from token-type indices in the case of sequence classification to maximum sequence cut-off, considering the added unique tokens of model-specific since most pre-shaped Transformer models have a maximum sequence length. Heads are usually added to pre-trained transformer models using transfer learning to adapt them to particular tasks. Each transformer model can be combined with one of the several ready-made heads with output compatible with common tasks. Furthermore, the \textit{Huggingface Model Hub} Fig.\ref{secondFig} facilitates the access and download of models, allowing researchers and developers to use it easily. It also provides community-based approaches that allow users to contribute to their own models or improve existing models. Using model hubs, developers can save time and effort using pre-trained models rather than starting from the beginning. \footnote{Transformers Model Hub https://huggingface.co/models}.

\subsection{Typical Large Pre-trained Language Model}

The table.\ref{tab:model} below shows the model since 2013\cite{Koutsomitropoulos_2021}. But in this paper, we focus only on BioGPT and BioBERT as these models are largely trained on biomedical text corpora with higher accuracy on biomedical regression tasks.
\begin{table}[!htb]
\center
\caption{The table shows the series of models used for language processing tasks.}
\label{tab:model}
\begin{tabular}{lllll}
\hline\noalign{\smallskip}
Model &   Released Year  &  Founders &  ANN  &   Vector per  \\
\noalign{\smallskip}
\hline 
\noalign{\smallskip}
Word2Vec & 2013 &  Mikolov & Shallow &  Word \\
Doc2Vec & 2014 & Mikolov  & Shallow  &  Word and Paragraph  \\
FastText  & 2016 &  Facebook  & - &  Syllable \\
 ELMo  & 2018 &  Allen Institute &   Deep  & Character  \\
BioBERT  & 2019 &   Google  &   Deep  &  Sub-word  \\
 BioGPT  & 2023 &   Microsoft  &   Deep  &  Sub-word  \\
GPT-4  & 2023 &   OpenAI  &   Deep  &  Sub-word \\
\hline
\end{tabular}
\end{table}

\textbf{BioGPT} A generative transformer language model to generate and extract biomedical texts, adopted GPT-2 as the basis model and pre-trained in the 15M PubMed abstract corpus. The BioGPT team carefully designed and investigated prompts and target sequence formats when applying BioGPT pre-trained to downstream tasks, including document classification tasks. They followed the same training/test separation as in transfer learning evaluation \cite{Peng_2019}. We use continuous embedding with length=1 as an indicator and format the label in the target sequence as previously described and fine-tuned GPT-2 medium and BioGPT for 20,000 steps with a maximum learning rate of $10^{-5}$ and 1000 warming steps with an F-1 score of 85.12 in HoC (the corpus of Hallmarks of Cancer)\cite{Kevin_2022}.

\textbf{BioBERT}
While BERT is a powerful language representation model that has been pre-trained on Wikipedia and BooksCorpus, it may struggle with extracting biomedical text due to the specific domain names and terminology used in this field. Peng and his team created BioBERT, a custom tool intended for use with PubMed abstracts and full-text articles from PubMed Central, in order to address this issue. They experimented with different pre-training structures and found that increasing the number of training steps improved the model performance. In BioBERT v1.0 (PubMed), the F1 score was 2.80 higher than that of other advanced models, demonstrating its effectiveness in extracting biomedical text\cite{Lee_2019}.

\section{Material and Methods}

\subsection{Prerequisite}
This section will discuss the algorithms behind the models, ontology retrieval, and the BioPython module.

\noindent\textbf{Algorithms}
Based on our scenario, we chose DistilBERT\cite{Sanh_2019} as our backbone model for further training. 
The model is inherited from transformer architecture, a new approach to addressing sequence-to-sequence tasks, with the added ability to handle long-range dependencies with ease. The transformer approach differs from conventional methods that utilise RNNs or convolution for sequence alignment. Instead, it relies exclusively on self-attention to generate representations of the input and output.

\begin{figure}[!htb]
\center
\includegraphics[scale=0.5]{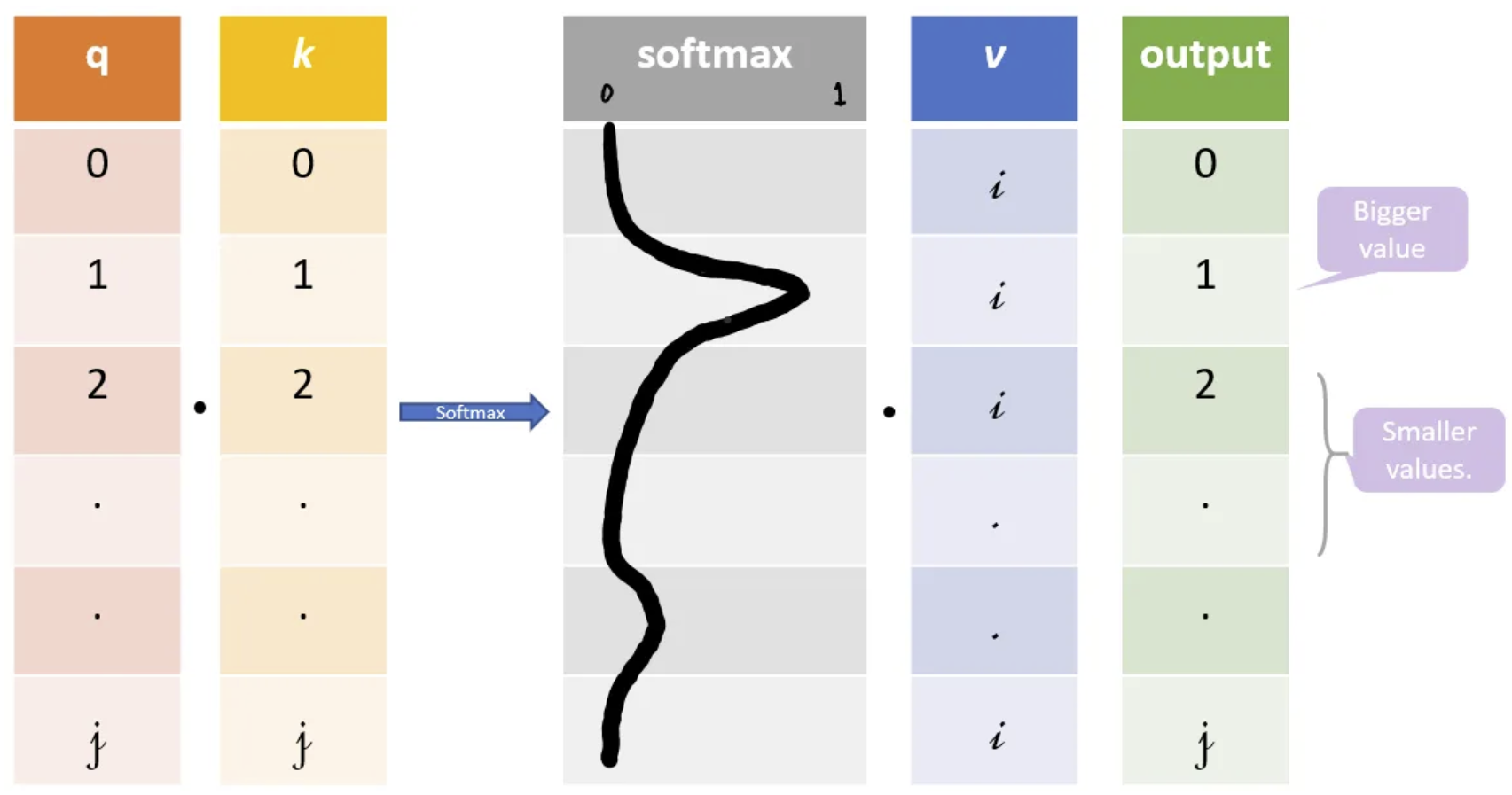}
\caption{The value vectors correspond to q, k, and i indices and are multiplied by the distribution, assigning greater importance to the more significant vectors. The figure adapted from illustrated transformer\cite{jalammar2018}}
\label{thirdFig}
\end{figure}

\textit{Self-attention} in our model allows us to focus on certain parts of the input sequence to predict the output words. The following code snippet maps the diagnosis based on human papillomavirus (HPV), messenger RNA (mRNA), and isolated systolic hypertension (ISH), indicating a strong correlation between these terms. When incorporating self-attention into the input sequence, it encompasses all words within the sequence, whereas, in the output sequence, it needs to be restricted to the words preceding a specific word for accurate attention. This prevents any information leakage during model training by hiding the words that appear after each step.
\begin{verbatim}
" HPV mRNA ISH contributes to the accurate diagnosis and grading 
ofCIN and has better specificity than IHC staining of p16. "

diagnosis => HPV mRNA ISH

\end{verbatim}
When processing text, we use the query vectors (1)
$$
q= X * Wq \eqno{(1)}
$$
key vectors (k)
$$
k= X * Wk \eqno{(2)}
$$
value vectors (3). 
$$
v= X * Wv \eqno{(3)}
$$
To obtain the final output of the transformer, we multiply the input vector (X) by the weight matrices we learnt during training. The query vector represents the current word, while the key vector acts as an index for the corresponding value vector, which contains information about the word. To find the most similar key vector for a given query vector, we perform a dot product and then apply a \textit{softmax} function to provide a distribution that emphasises the most relevant key-value pairs. We then multiply the value vectors by this distribution, placing higher weights on the more essential vectors. The size of these vectors is often referred to as the "hidden size" in various implementations. Fig.\ref{thirdFig} illustrates that the transformer will pick up the most robust related parameter\cite{Kulshrestha2023}.

\begin{figure}[!htb]
\center
\includegraphics[scale=0.5]{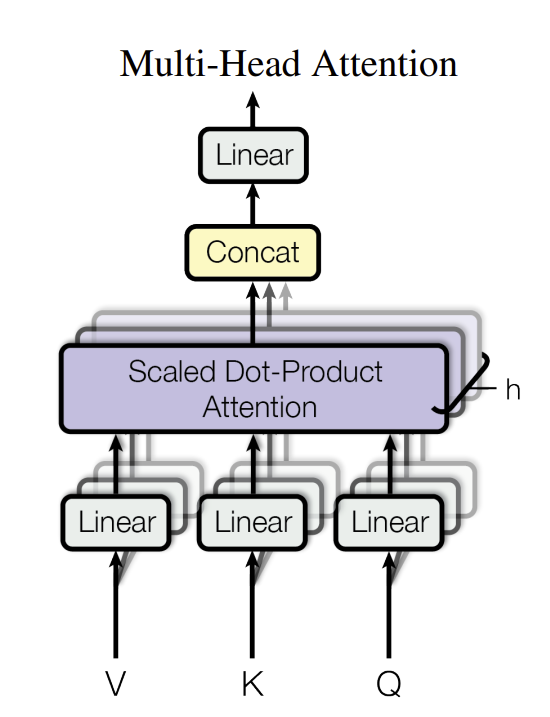}
\caption{Multi-Head Attention involves multiple attention layers that operate simultaneously in parallel. The figure adapted from illustrated transformer\cite{jalammar2018}}
\label{forthFig}
\end{figure}

By calculating the value of the key and the query, the model uses the softmax function to ensure that all possibilities of the words are positive.
$$
\operatorname{Attention}(Q, K, V)=\operatorname{softmax}\left(\frac{Q K^T}{\sqrt{d_k}}\right)V
$$
When combined and projected, these values combine to form a final result, as shown in Fig.\ref{forthFig}. The use of multi-head attention allows the model to respond simultaneously to information from various representation subspaces in different positions, which is impossible for a single attention head due to the average effect.\cite{Vaswani_2017}
$$
\begin{aligned}
\operatorname{MultiHead}(Q, K, V) 
& =\operatorname{Concat}\left(\operatorname{head}_1, \ldots, \text { head }_{\mathrm{h}}\right) W^O \\
\text { where head }_{\mathrm{i}} 
& =\operatorname{Attention}\left(Q W_i^Q, K W_i^K, V W_i^V\right)
\end{aligned}
$$
The Multi-Head Attention method utilizes several attention layers that work together in parallel. The parameter matrices used for the projections include
$W_i^Q\in\mathbb{R}^{d_{model} \times d_k}, W_i^K \in \mathbb{R}^{d_{model}\times d_k}, W_i^V \in \mathbb{R}^{d_{model} \times d_v}$ and $W^O \in\mathbb{R}^{h d_v \times d_{model}}$.
To ensure optimal performance, they used $h = 8$ attention layers or heads, each with a reduced dimensionality of $dk = dv = dmodel/h = 64$. Due to this reduction in dimensionality, the computational cost is similar to that of single-head attention with full dimensionality.

\noindent\textbf{Ontology Retrieval}
A new version of NCBO Ontology Recommender, called Ontology Recommender 2.0, uses a unique approach to evaluate the relevance of an ontology to biomedical text data based on four criteria described in the table\cite{Martínez-Romero_2016}. 

\begin{table}
\center
\label{table2}
\caption{The table includes ontology coverage of input data, acceptance among the biomedical community, detail level in its input data class, and specialization in the field of input data.}
\addtolength\tabcolsep{5pt}
\begin{tabular}{llll}
\hline\noalign{\smallskip}
Criteria 1 &   Criteria 2  &  Criteria 3 &  Criteria 4  \\
\noalign{\smallskip}
\hline 
\noalign{\smallskip}
input data
& biomedical verification 
 & all details & specific domain \\
\hline
\end{tabular}
\end{table}
According to Martínez-Romero and her associates' assessment, the improved promoter provided better suggestions than the original method, offering more coverage of input data, more detailed concept information, increased specialization of the input data domain, and a higher level of acceptance and use within the community. In addition, it offers users more explanation and suggestions for the use of individual ontologies and ontologies groups together, while allowing customization to adapt to various ontology recommendation scenarios\cite{Martínez-Romero_2016}.
 NCBO Ontology Recommender 2.0 uses a separate recommendation approach to evaluate the relevance of ontology to biomedical text data, unlike the NCBO annotation method. The new process is driven by annotations, but the assessment is based on a scoring system that uses specific equations to calculate the points for each annotation\cite{Martínez-Romero_2016}. 
 \begin{gather}
annotationScore2(a)=\left( a nnotationTypeScore(a)+ multiWordScore(a)\right)\\
 * annotatedWords(a)
\end{gather}
and
$$
annotationTypeScore(a)=\left\{\begin{array}{c}\hfill 10\; if\; annotationType= PREF\hfill \\ {}\hfill 5\; if\; annotationType= S Y N\hfill \end{array}\right.
$$
$$
multiWordScore(a)=\left\{\begin{array}{c}\hfill 3\; if\; annotatedWords(a)>1\hfill \\ {}\hfill 0\; otherwise\hfill \end{array}\right.
$$

In our project, we need to extract both experimental designs and laboratory techniques from the graph ontology for the future classification task, as shown in the figures below. 

\begin{figure}[!htb]
    \centering
    \subfloat[\centering experimental design]{{\includegraphics[scale=0.5]{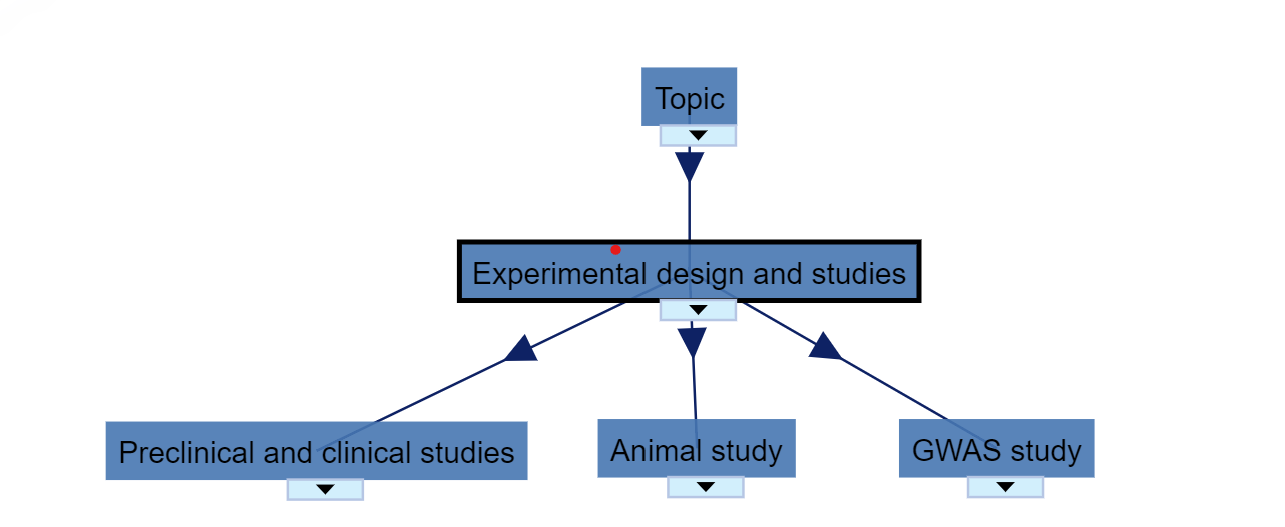} }}%
    \qquad
    \subfloat[\centering laboratory technique]{{\includegraphics[scale=0.5]{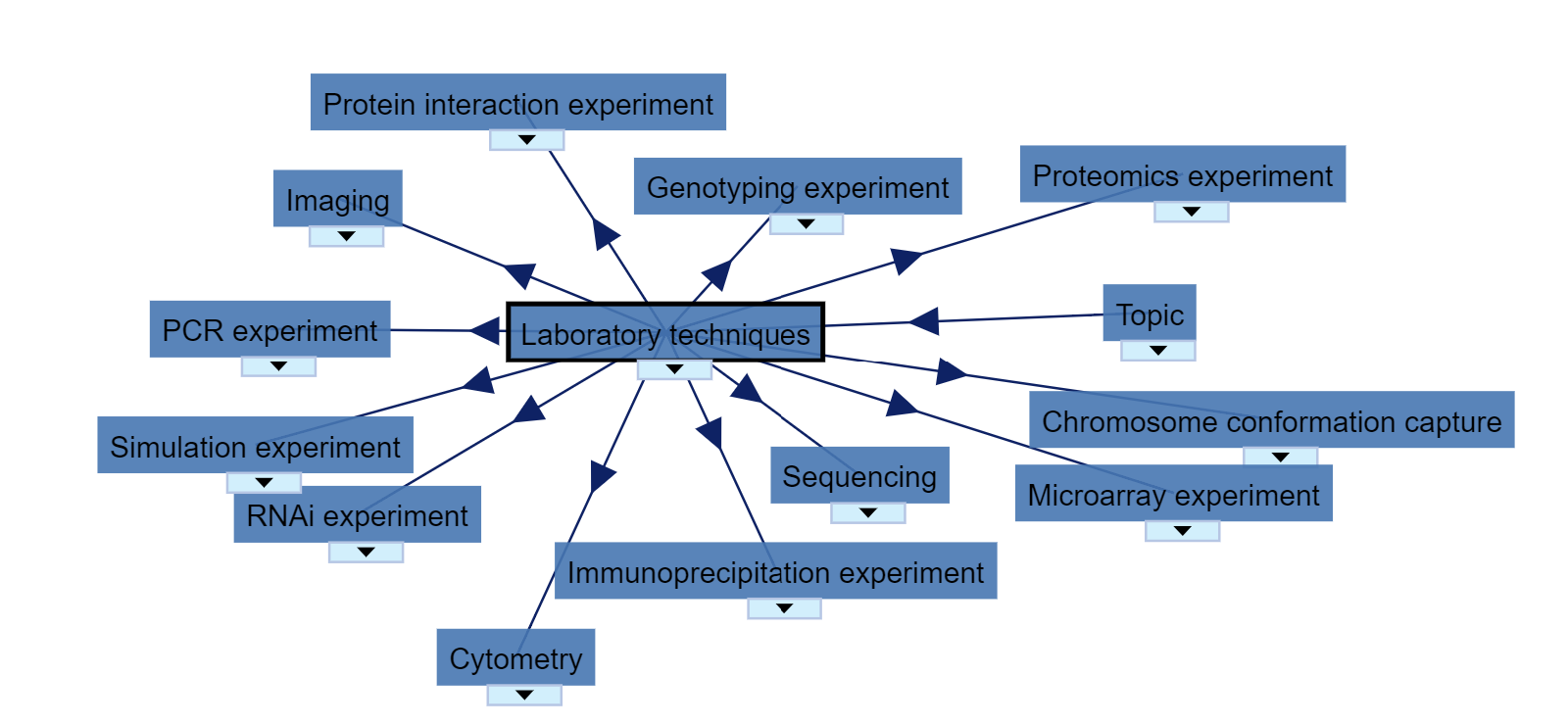} }}%
    \caption{(a) on the top shows three sub-categories of experimental design. (b) on the bottom illustrates the 12 techniques of the topic. more details: \emph{https://bioportal.bioontology.org/ontologies/EDAM/}}%
    \label{duoFig}%
\end{figure}
As a result, Python's proto\cite{martin_larralde_2023_7814219}
library enables the manipulation and examination of Ontology Web Language (OWL) and Open Biological and Biomedical Ontology (Open Biological and Biomedical Ontologies) format files. By using this tool, it is super useful for analysing and analyzing ontologies representing knowledge in different domains.
In the code snippet below, we loop the sets of EDAM ontology and extract all the subsets of both experimental designs and laboratory techniques. 
\begin{figure}[!htb]
    \center
    \includegraphics[scale=0.5]{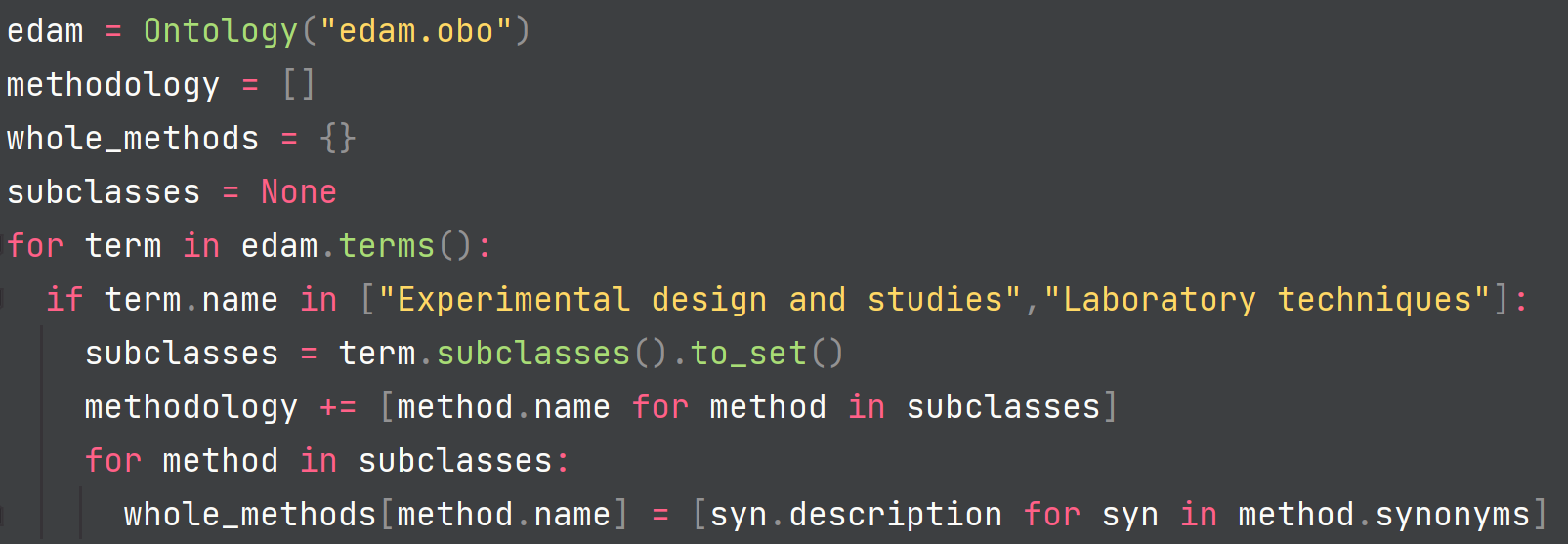}
    \caption{The ontology extraction code fragment}
    \label{code1}
\end{figure}\\

\noindent\textbf{Biopython Setup}
The modules within Bio-python\cite{Cock_2009} have expanded significantly and are designed for programmers in the fields of computational biology or bioinformatics to use in our scripts or integrate into our own data flow pipeline. In the bio-python submodule - Entrez\cite{Maglott_2004}, a gene database, we can easily set up the biomedical literature retrieval tool by setting the environment with our email account.
\begin{figure}[!htb]
    \center
    \includegraphics[scale=0.5]{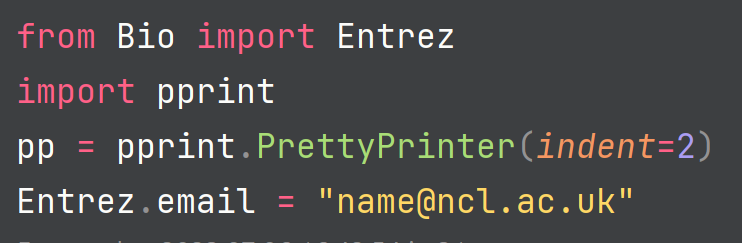}
    \caption{Bio-python environment setup}
    \label{code2}
\end{figure}\\
\subsection{Data Acquisition}
The pmids\footnote{refers to a unique id list of articles} was provided by my supervisor, consisting of over 30,000 scholarly essays relating to disease-gene association studies. Our starting point was to retrieve all abstracts and full text of scientific biomedical articles. Our primary resource was the two articles BioBert\cite{Lee_2019} and BioGPT\cite{Luo_2022}, which describe NLP methods in biomedical-specific areas. Entrez provides search and summary functions that can be easily used to get the abstract and details of the target article. However, we only pulled the paper's method and result sections in the full-text extraction task. In the Results section, there is a significant difference in the outcomes of the experiments. It can be found by the characteristic of the data presented in a logical order, with subheadings used to separate the results of different experiments. Similarly, the method section tends to describe which methods have been used in the experience or in the overall architecture. Therefore, the method and result sections accurately reflect the main content of the paper.
We searched for abstracts relevant to methodology research and PubMed journals in sub-areas of biomedicine, such as imaging, RNA sequence, and cytometry, using a set of representative search terms for each of the methods provided by NCBO ontology\cite{Martínez-Romero_2016}. The search terms were chosen based on domain-specific techniques, initially setting up for 42 terms, and we used subcategory terms, the methodology's synonyms, to expand the labels toward 188. 
Articles without abstracts were discarded. Meanwhile, we narrowed the provided data set down to 3200 articles recognising the method used in the articles. In Entrez API, the results were limited to 10,000 in each investigation to ensure that no large amount of data was returned at one time\cite{Baker_2016}. Therefore, the number of abstracts retrieved per method ranged from more than a hundred to several thousand, and we only downloaded the XML format. Due to the limitation of the PubMed dataset, we can only extract the full text of the article by accessing the PubMed subset provided by BioC\cite{Comeau_2013}. The subsets enable us to focus on a specific section of the database, and, to help us with the search, PubMed elaborates on two types of subsets available, namely processing status subsets and a subject subset for AID, and the corresponding subset tags to use when performing your search. Consequently, we combined Entrez keyword search and BioC full-text retrieval tool to retrieve the final data set including abstracts, methods, and results. 
\begin{table}
\center
\label{table3}
\caption{Hallmark and Search terms(Source: table credit original).}
\begin{tabular}{ll}
\hline\noalign{\smallskip}
Hallmark  &   Search Terms  \\
\noalign{\smallskip}
\hline 
\noalign{\smallskip}
Tomography & CT, PET, Computed tomography, TDM\\
\noalign{\smallskip}
Imaging & \makecell[l]{Microscopy imaging, Diffraction experiment,\\ Optical super-resolution microscopy,\\ Photonic microscopy}\\
\noalign{\smallskip}
RNA immunoprecipitation & RIP, PAR-CLIP, CLIP, CLIP-seq, HITS-CLIP, iCLIP\\
\noalign{\smallskip}
NMR & \makecell[l]{Rotational Frame Nuclear Overhauser Effect Spectroscopy\\,Nuclear magnetic resonance spectroscopy}\\\noalign{\smallskip}
Neutron diffraction&\makecell[l]{Neutron diffraction experiment, Neutron microscopy\\, Elastic neutron scattering}\\\noalign{\smallskip}
X-ray diffraction&Crystallography, X-ray crystallography, X-ray microscopy\\
\hline
\end{tabular}
\end{table}

\begin{figure}[!htb]
    \center
    \includegraphics[scale=0.5]{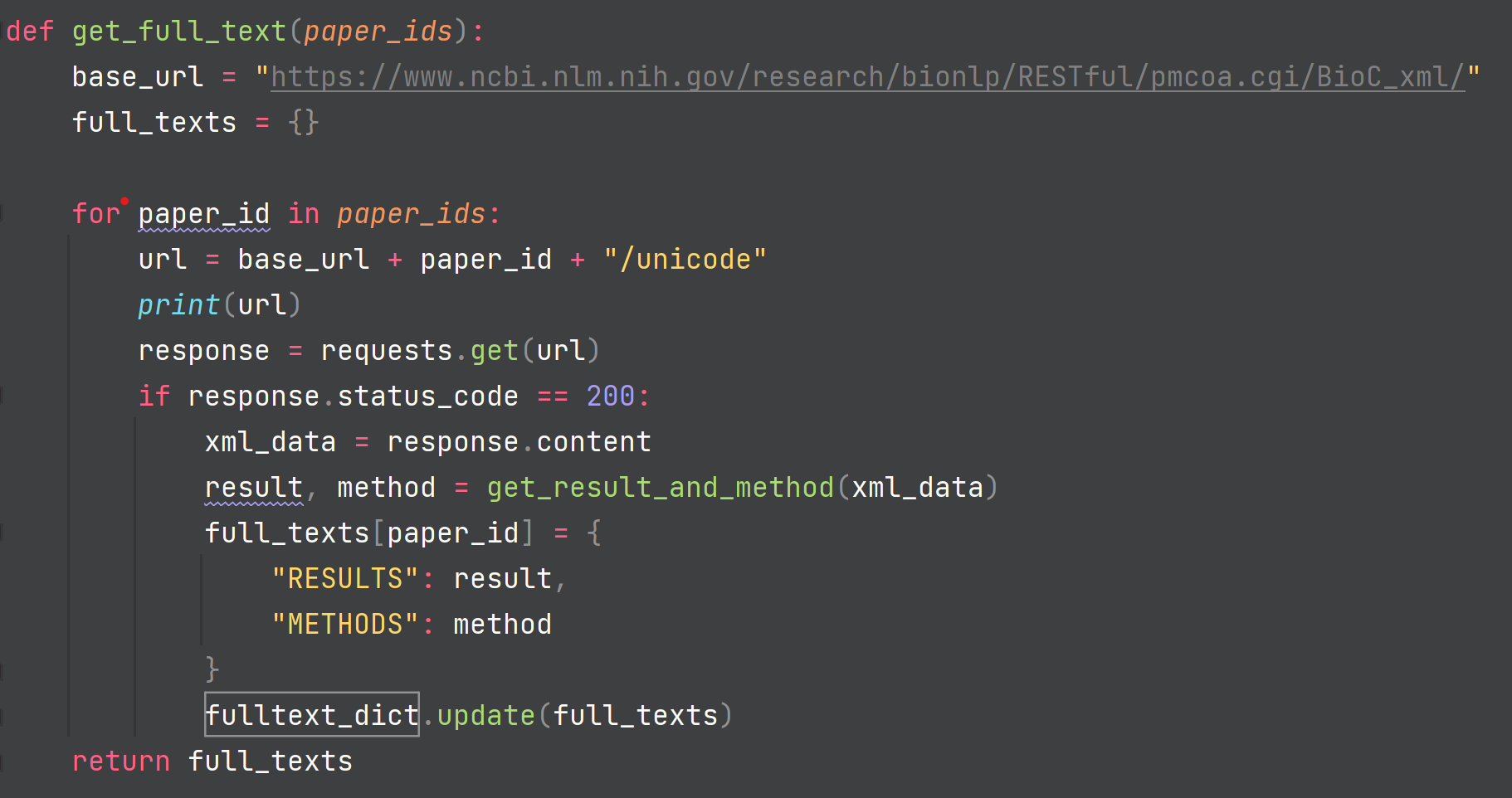}
    \caption{Full text retrieval tool}
    \label{code3}
\end{figure}

\subsection{Pre-processing}
To train the model by the extracted abstracts and full-text data, we need to label each text to ensure that our model is under supervised learning. By doing this, we marked each abstract or full text with a different methodology. One content might have one or multiple methods marked depending on whether the content has numerous labels defined in the ontology. Even if the HuggingFace pipeline provides the trainer function to train the raw Python dictionary-type data, we need to glimpse the overall data scattering. Therefore, we first converted the Python dictionary into a pandas data frame and, second, used a filter function to map different types of content with numeric numbers. Hence, the matrix can be constructed into a sparse matrix (the row refers to the unique id of the paper, and the column refers to a different label).To make raw data understandable, we implemented data pre-processing techniques through a data flow pipeline classifier. By eliminating less important data characteristics and enhancing accuracy, instead of using the traditional NLP toolkit NLTK, we utilised a novel hugging face pre-processing approach that imports the pre-trained tokenizer to format the input sequence.
To establish the cluster view, SciPy\cite{2020SciPy-NMeth} provides a dendrogram function to visualise how closely each cluster is related. SciPy first links different rows by binary label matrix and plots the distance between each group. In the case of hierarchical clustering, the algorithm treats each data point as an independent cluster. Then it combines a cluster based on its similarity until all data points are a single cluster. When the cluster merges, a new cluster is created and gives a different identifier, such as index or branch, to differentiate it from other clusters.
\begin{figure}[!htb]
\center
\includegraphics[scale=0.4]{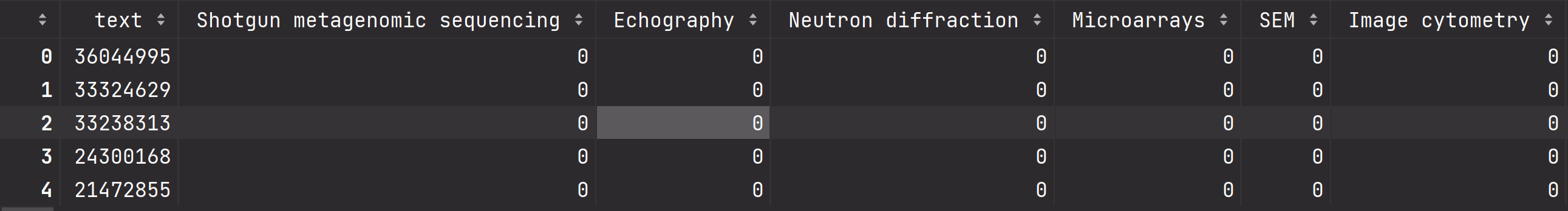}
\caption{First column refers to the unique id of the article, the rest of the columns stand for different biomedical terms defined in NCBO\cite{Martínez-Romero_2016} and data are binary numbers.}
\label{matrix}
\end{figure}
\begin{figure}[!htb]
\center
\includegraphics[scale=0.4]{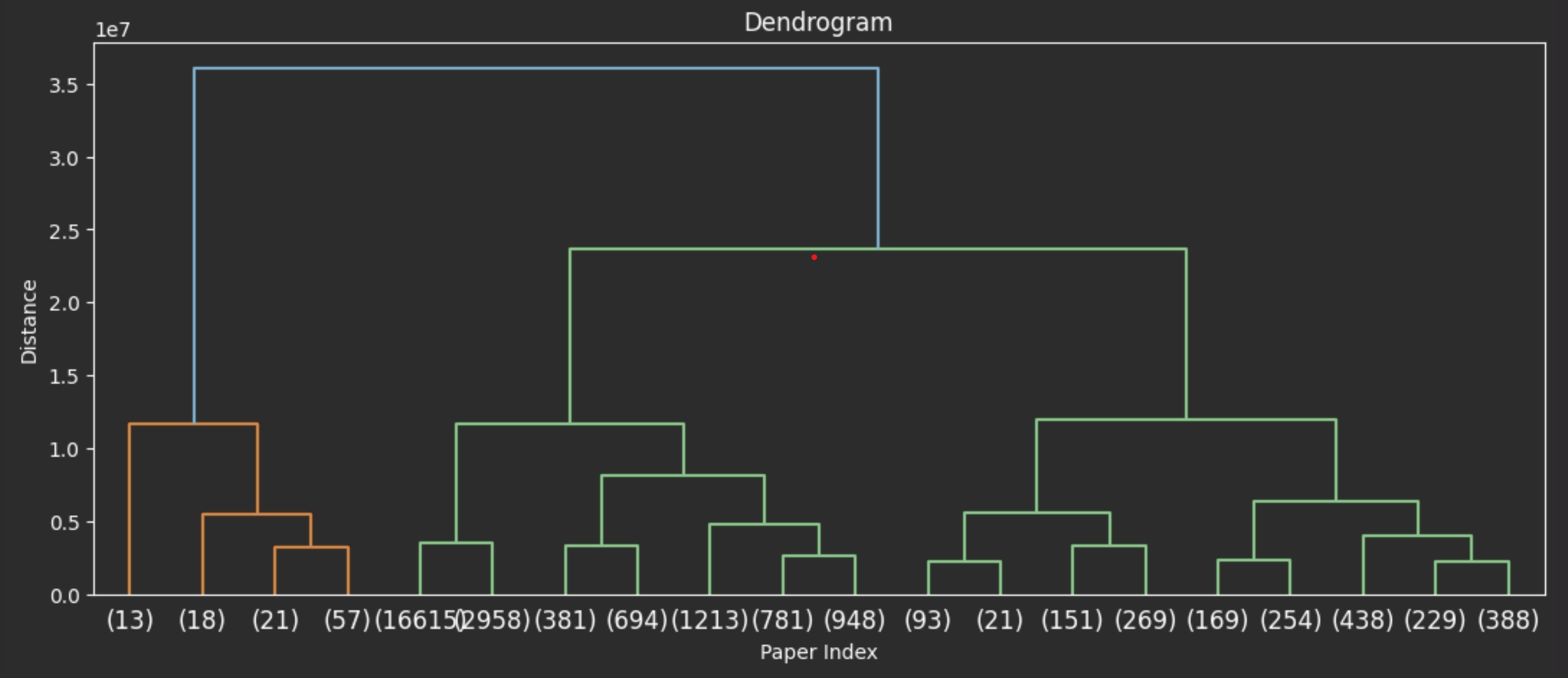}
\caption{X-axis shows the index of each article, and y-axis shows the similarity between each data point. Also the entire dendrogram could be super large so we truncated the branches to make the figure can contain the drawn dendrogram.}
\label{dendrogram}
\end{figure}\\
The text in the dataset underwent several stages, including lowercasing, tokenization, cleaning, and lemmatization, before entering the learning process. In fact, we used a smaller size of the uncased-based tokenizer to save more space for GPU usage; therefore, lowercase is not necessary. The tokenizer breaks down text into smaller units known as tokens, which is referred to as tokenization, which is vital in preparing text data for use in natural language processing tasks in transformer models. It helps our model to comprehend the structure of the text and handle language variations. Finally, lemmatisation removes inflection. Once we finished the previous step, we split the dataset into two parts - one for "Learning" and the other reserved for "Test". To achieve this, we assigned 80\% of the data to the training set, while the remaining 20\% was allocated for testing.
\subsection{Model Selection}
BERT model is unique in that it is bidirectional, whereas the previous models were unilateral and read text in a specific direction\cite{Peng_2019}. BERT consists of multiple layers that form a "transformer" that learns contextual relationships between words in the text. Transformers strive to analyse the terms of complex questions to link them better to understand the phrase's semantics and overall meaning. One way to enhance the speed and efficiency of transformers is to utilise TPU clouds, which are integrated circuits that accelerate their workload. Compared to modern GPUs or CPUs, the TPU is approximately 15-30 times faster on average, while its TOPS/Watt is roughly 30-80 times faster\cite{DBLP:journals/corr/JouppiYPPABBBBB17}. In our proposal, due to the limitation of the local training environment and the high cost of the TPU unit, we used the DistilBERT architecture for the multi-label classification of biomedical literature and an input text consisting of an abstract or full details of the article\cite{app12062891}. To rule out the optimised model, we need to set the model problem-solving type as multi-label classification and specify the labels by using the pre-processed label number. 
\begin{figure}[!htb]
\center
\includegraphics[scale=0.4]{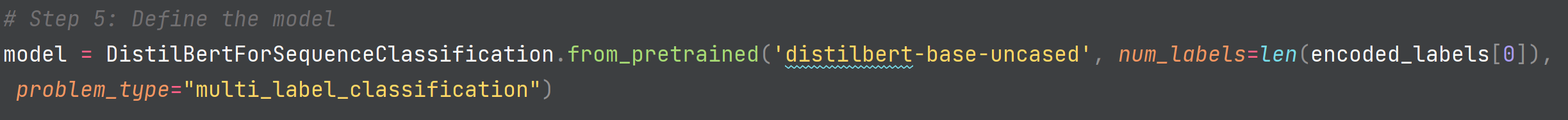}
\caption{Defined DistilBERT model}
\label{model}
\end{figure}

\subsection{Fine-tune}
This section presents a guide on modifying the pre-trained DistilBERT for biomedical literature methodology classification in downstream tasks. All task inputs are sequences padded to the maximum length of sentence - 512 tokens. As DistilBERT is pre-trained on a large corpus of natural language, we transformed the labels into natural language sequences instead of the structured format that was used in previous BERT models, which differs from the tokens used in \cite{gd2022sequencetosequence,plh2021rebel}. We fine-tuned DistilBERT based on uncased for ten epochs, using an AdamW optimiser with a maximum learning rate of $10^{-5}$ and BCEWithLogitsLoss as the loss function (a combination of BCEloss and sigmoid). Since we trained for multi-label prediction, we customised the output logits using raw Pytorch for further predictions.
\subsection{Challenges}
\textbf{Scale of Data} Managing a large number of documents can be a daunting task, especially when some of them are difficult to understand or no longer relevant. Sorting through more than 20,000 articles to identify which can be discarded or archived can be time-consuming and challenging. Furthermore, dealing with arcane documents, for example, an animal study of Hong and his colleagues\cite{Hong_2022}, can add another layer of complexity to the process. It is essential to scale the data to 20,000 in place to effectively manage the training data shortage. \\
\textbf{Preprocessing} When dealing with raw Python dictionary data, the computer may struggle to interpret the data due to its unstructured nature. This can make it difficult to extract meaningful insights and patterns from the data. Furthermore, the data may contain errors or inconsistencies that need to be addressed before they can be used effectively in a model. These challenges require careful pre-processing and cleaning of the data to ensure that it is properly formatted and ready for analysis. By taking the time to address these challenges, we can ensure that our models are accurate and effective, leading to better results and more efficient use of our resources. \\
\textbf{Hardware limitation} It has been quite a challenge for us to work around the limitations imposed by our GPU. We have found that our choices of models are extremely limited, which has made it difficult to achieve the level of performance we are looking for. 
\section{Results and Discussion}

\textbf{Evaluation}
In this section, we outline some indicators for assessing the quality of the model. To evaluate the performance of the classifier, four types of elements are classified for the desired class. True positive (TP), false positive (FP), false negative (TN), and false negative (FN)\cite{Heydarian_2022}. TP represents the positive class predicted by the model correctly, while FP indicates the positive class predicted by the model incorrectly. TN represents the false class that the model correctly predicts and FN represents the false class that the model incorrectly predicts. We will now introduce the evaluation criteria for measuring the performance of various Deep Learning(DL) models. Our assessment is based on five different measures, accuracy, accuracy, recall, hamming loss\cite{Wu_2020}, and F1 score\cite{Heydarian_2022}. These evaluation metrics are defined as follows:
\begin{enumerate}
    \item Accuracy is a measure of the classifier's overall accuracy, defined as the ratio of the sample to the sample number correctly predicted with regard to the section of accuracy in scikit-learn\cite{scikit-learn}. 
    $$
    \text{Accuracy} = \frac{TP + TN}{TP + TN + FP + FN}
    $$
    \item The recall, also known as sensitivity or real positive rate (TPR), measures the classifier's ability to identify positive samples correctly. It refers to the proportion of actual positive cases that are correctly identified by a diagnostic test or machine learning algorithm by viewing the section of recall in scikit-learn\cite{scikit-learn}.
    $$
    \text{Recall} = \frac{TP}{TP + FN}
    $$
    \item  The precision measure measures the classifier's ability to correctly identify a positive sample from a positive sample. This is the ratio of the sample of the correctly predicted positive sample to the total number of the predicted positive sample considering the precision section in scikit-learn\cite{scikit-learn}. 
    $$
    \text{Precision} = \frac{TP}{TP + FP}
    $$
    \item The F-1 Score is a metric that balances the precision and recall measures, providing a more accurate evaluation when dealing with imbalanced datasets that have uneven class distribution regarding the section of F-1 score in scikit-learn\cite{scikit-learn}.
    $$
    \text{F1} = 2 \cdot \frac{\text{Precision} \cdot \text{Recall}}{\text{Precision} + \text{Recall}}
    $$
    \item In the case of multi-label classification, Hamming Loss is a measurement that evaluates the proportion of labels that are incorrectly classified with the total number of labels in a dataset\cite{Wu_2020}.
    \begin{align*}
        \text{Hamming Loss} = \frac{1}{n \cdot L} \sum_{i=1}^{n} \text{hamming}(i)\\
        \text{hamming}(i) = \frac{\text{number of misclassified labels in sample } i}{L}
    \end{align*}
\end{enumerate}

\subsection{Experimental Results}

We delved into DL and experimented with training constraints before tackling the multi-label text classification task using the DistilBERT-base model\cite{Sanh_2019}. Our approach involved automatically collecting biomedical papers related to all methods extracted from the NCBO ontology\cite{Martínez-Romero_2016}. We proceeded to train on our own, training the data loader and evaluating the data loader, which consisted of 20568 and 5143 biomedical scientific papers distributed evenly across 188 different categories. To tailor the existing DistilBERT model to our specific requirements, we made modifications and trained it on our digitised labels. To enhance learning costs, we use the DistilBERT unboxing base model together with the RTX 2070 Super Q Design Graphics Processing Unit (GPU). We also refined training hyperparameters, such as batch size, maximum length, learning rate, seed value, and period number, while closely monitoring validation and training losses. We selected the optimal model based on the validation set.

\begin{figure}[!htb]
\center
\includegraphics[scale=0.5]{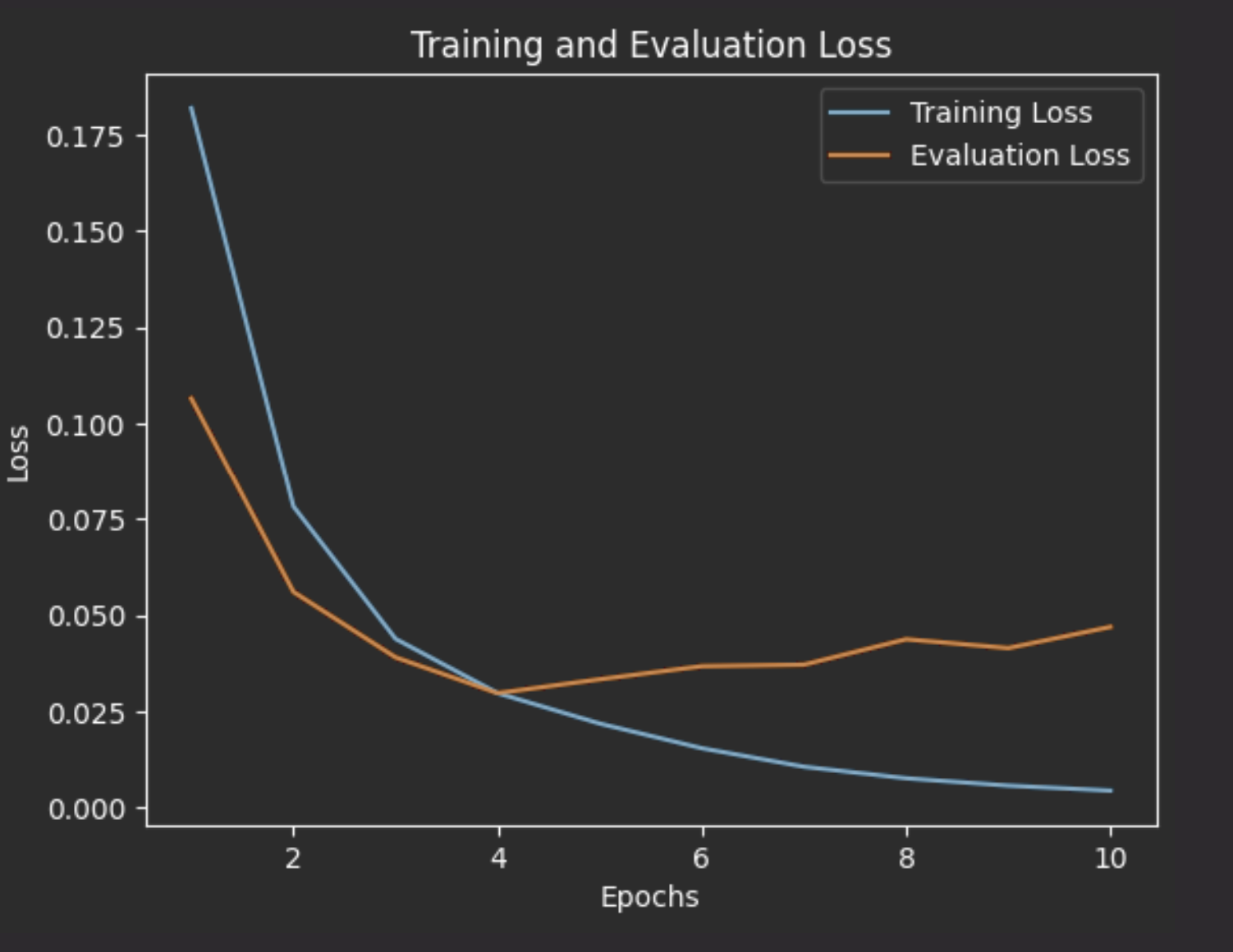}
\caption{The graph respectively illustrates the training loss and validation loss during the 10 epochs. As we can see at the time of the 4th epoch, the validation and training loss are the same and the model was just in fit.}
\label{loss}
\end{figure}

Using our fine-tuned model, we successfully predicted the multi-label method by feeding the result section of an unseen biomedical article. We chose pre-extracted full-text articles and picked out a single result by specifying the article's unique id (19030899). We transformed the article into tokenised format and evaluated the model output through tokenizer input and attention mask. The result showed four method labels for this article, demonstrating our model's effectiveness in accurately classifying multi-label samples.

\begin{figure}[!htb]
\center
\includegraphics[scale=0.5]{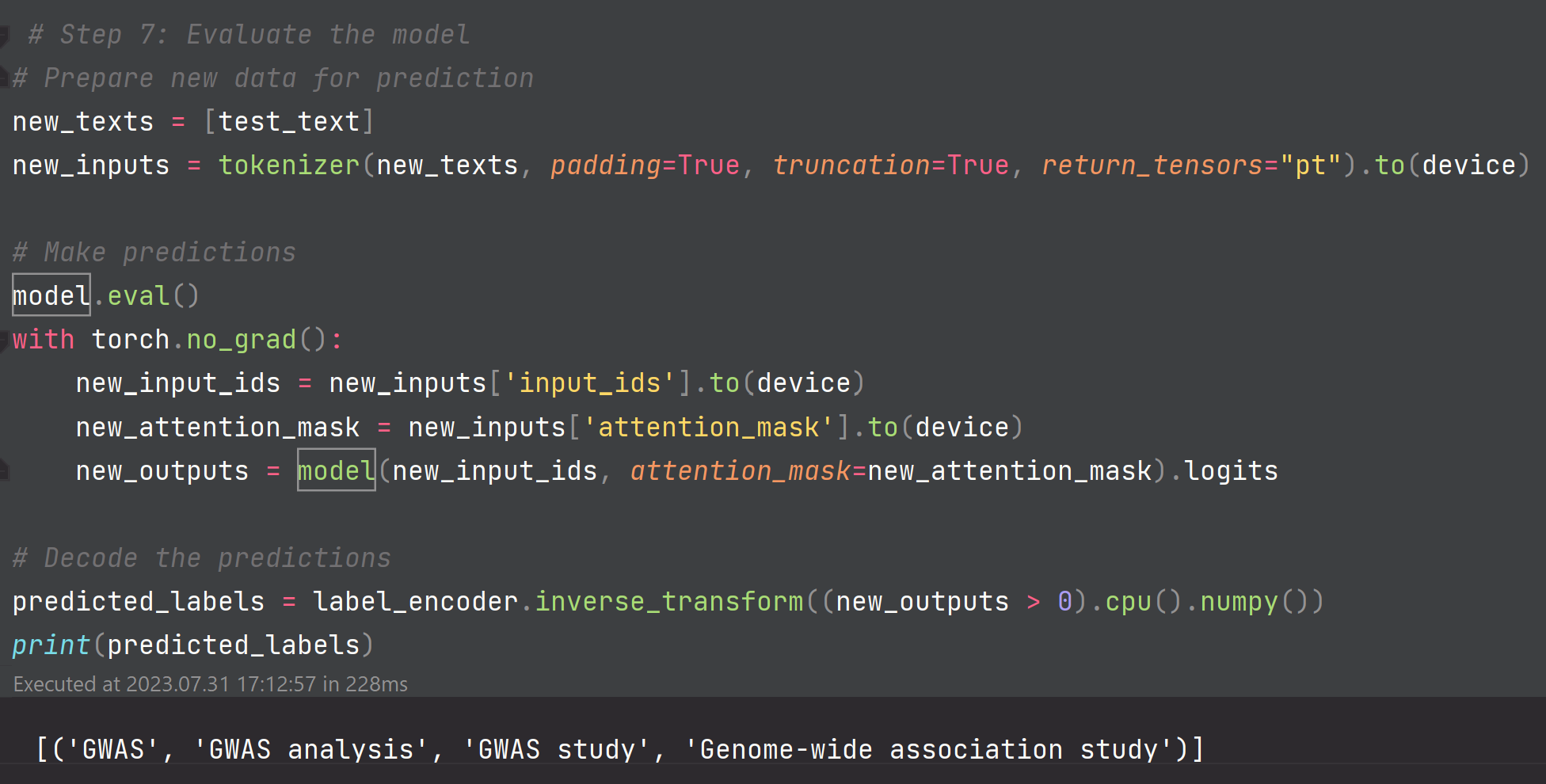}
\caption{The graph shows the output labels of the input text.}
\label{prediction}
\end{figure}

\subsection{Discussion}

\begin{table}[!htb]
\center
\caption{The table shows the marks of computed metrics. The left table model was trained by abstracts, and the right table model was trained by abstracts plus full texts.}
\label{score}
\begin{tabular}{ll}
\hline\noalign{\smallskip}
Evaluation Name &   Score  \\
\noalign{\smallskip}
\hline 
\noalign{\smallskip}
Accuracy& 0.7515\\
Precision& 0.7608\\
Recall& 0.7584\\
F1-score& 0.7578\\
Haming-loss& 0.0103\\
\hline
\end{tabular}
\quad
\begin{tabular}{ll}
\hline\noalign{\smallskip}
Evaluation Name &   Score  \\
\noalign{\smallskip}
\hline 
\noalign{\smallskip}
Accuracy& 0.7985\\
Precision& 0.72468\\
Recall& 0.7845\\
F1-score& 0.7458\\
Haming-loss& 0.0126\\
\hline
\end{tabular}
\end{table}

\section{Conclusion}
In this project, we have successfully implemented a multi-label classification model using the powerful DistilBERT pre-trained model and have developed our model of classification of biomedical literature that focuses only on methods classification of biomedical literature. Using DistilBert as a support model and pre-trained on a corpus of 32,000 abstracts and complete text articles, our results were impressive and surpassed those of traditional literature classification methods by using RNN or Wiki. Our aim is to integrate this highly specialised and specific model into different research industries, especially in the field of research in the biomedical literature. In the future, we plan to fine-tune the pre-trained language models using the larger text corpus through the masked language model and expand the model to adapt comprehensive methods. 

\textbf{Acknowledgments}
I would like to express my sincere gratitude to Professor Katherine James for her guidance and support throughout my individual coursework. Her expertise and insights have been invaluable in helping me navigate the complex topics and challenges presented in this project. Overall, this challenging project enriches my research interest in the fields of DL and NLP, and I hope to put more effort into my DL research career.


\end{document}